\documentclass[10pt,twocolumn,letterpaper]{article}

\usepackage{cvpr}
\usepackage{times}
\usepackage{epsfig}
\usepackage{graphicx}
\usepackage{amsmath}
\usepackage{amssymb}

\usepackage{epstopdf}
\usepackage{cite}

\usepackage{makecell}
\usepackage{multirow}
\usepackage{enumerate}
\newcommand{\PreserveBackslash}[1]{\let\temp=\\#1\let\\=\temp}
\newcolumntype{C}[1]{>{\PreserveBackslash\centering}p{#1}}
\newcolumntype{R}[1]{>{\PreserveBackslash\raggedleft}p{#1}}
\newcolumntype{L}[1]{>{\PreserveBackslash\raggedright}p{#1}}
\usepackage[pagebackref=true,breaklinks=true,letterpaper=true,colorlinks,bookmarks=false]{hyperref}

\newcommand{\tabincell}[2]{\begin{tabular}{@{}#1@{}}#2\end{tabular}}

\cvprfinalcopy 

\def\cvprPaperID{6573} 

\ifcvprfinal\fi
\begin{document}

\title{TracKlinic: Diagnosis of Challenge Factors in Visual Tracking}

\author{Heng Fan$^{1}$ \;\;\;\;\; Fan Yang$^{1}$ \;\;\;\;\; Peng Chu$^{1}$ \;\;\;\;\; Lin Yuan$^{2}$ \;\;\;\;\; Haibin Ling$^{3}$\\
	$^{1}$Department of Computer and Information Sciences, Temple University, Philadelphia, PA USA\\
	$^{2}$Amazon Web Services, Palo Alto, CA USA\\
	$^{3}$Department of Computer Science, Stony Brook University, Stony Brook, NY, USA\\
	{\tt\small \{hengfan,fyang,pchu\}@temple.edu \;\; lnyuan@amazon.com\;\; hling@cs.stonybrook.edu}
}

\maketitle

\begin{abstract}
   Generic visual tracking is difficult due to many challenge factors (\eg, occlusion, blur, etc.). Each of these factors may cause serious problems for a tracking algorithm, and when they work together can make things even more complicated.  
   Despite a great amount of efforts devoted to understanding the behavior of tracking algorithms, reliable and quantifiable ways for studying the per factor tracking behavior remain barely available. Addressing this issue, in this paper we contribute to the community a tracking diagnosis toolkit, \emph{TracKlinic}, for diagnosis of challenge factors of tracking algorithms. 
   
   TracKlinic consists of two novel components focusing on the data and analysis aspects, respectively. For the data component, we carefully prepare a set of 2,390 annotated videos, each involving one and only one major challenge factor. When analyzing an algorithm for a specific challenge factor, such \emph{one-factor-per-sequence} rule greatly inhibits the disturbance from other factors and consequently leads to more faithful analysis. For the analysis component, given the tracking results on all sequences, it investigates the behavior of the tracker under each individual factor and generates the report automatically. With TracKlinic, a thorough study is conducted on ten state-of-the-art trackers on nine challenge factors (including two compound ones). The results suggest that, heavy \emph{shape variation} and \emph{occlusion} are the two most challenging factors faced by most trackers. Besides, \emph{out-of-view}, though does not happen frequently, is often fatal. By sharing TracKlinic, we expect to make it much easier for diagnosing tracking algorithms, and to thus facilitate developing better ones.
\end{abstract}

\section{Introduction}

\begin{figure*}[!t]
	\centering
	\includegraphics[width=1\linewidth]{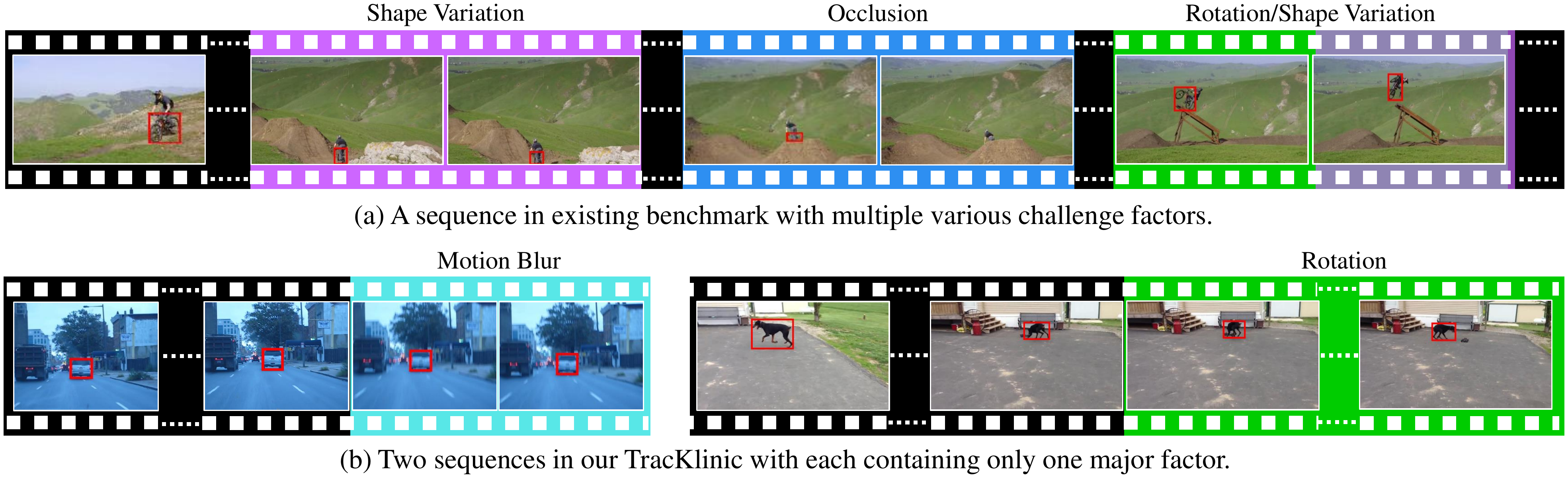}\\
	\caption{Comparison between existing tracking benchmark LaSOT~\cite{fan2019lasot} and our diagnosis benchmark TracKlinic. Note that, the {\it per frame} factor annotations in image (a) are labeled by us, and the original benchmark only provides global factor annotations. Best viewed in color. }
	\label{fig1}
\end{figure*}

As one of the most fundamental problems in computer vision, visual tracking has been studied for several decades. Despite considerable progress in recent years, object tracking remains difficult due to many challenge factors\footnote{Note that, \textit{challenge factor} is often called {\it attribute} as well~\cite{wu2013online}.} such as {\it occlusion}, {\it rotation}, {\it background clutter}, {\it out-of-view}, {\it motion blur}, etc. Any one of these factors may cause severe variation in target shape and/or appearance in a video, resulting in tracker failures. Moreover, multiple factors may occur at the same time, making things even worse.

Numerous approaches have been proposed to tackle the aforementioned challenge factors. To validate effectiveness, evaluating and comparing these trackers on datasets~\cite{wu2013online,wu2015object,fan2019lasot,muller2018trackingnet,kristan2018sixth,kristan2017visual,liang2015encoding,li2015nus,valmadre2018long,kiani2017need,smeulders2013visual,mueller2016benchmark} is an important section. Doubtlessly, these benchmarks greatly advance the tracking community. Nevertheless, if used directly, they are incapable of faithfully investigating tracker behaviors on specific factors, as described below.

In existing datasets, a sequence is typically involved with multiple challenge factors (see Fig.~\ref{fig1} (a) for an example), and these factors are usually annotated for the entire sequence instead of for individual frames. In such case, it is hard to {\it faithfully} examine a tracker on a specific factor due to disturbances from other ones. Likewise, {\it reliably} comparing different trackers with respect to a factor is hard, because the trackers may fail due to other factors instead of the one to assess. For instance, when comparing two trackers $\mathcal{T}_{A}$ and $\mathcal{T}_{B}$ on a sequence with {\it background clutter} and {\it occlusion}, assuming it only contains these two factors and {\it background clutter} occurs earlier, if $\mathcal{T}_{A}$ outperforms $\mathcal{T}_{B}$, can we conclude that $\mathcal{T}_{A}$ performs better than $\mathcal{T}_{B}$ in dealing with {\it occlusion}? Obviously, drawing such a conclusion would be arbitrary since $\mathcal{T}_{B}$ may fail where {\it background clutter} happens and it never has a chance to demonstrate its ability in handling the subsequent {\it occlusion}.

To conduct reliable and quantifiable {\it per factor} analysis for tracking algorithms, a new type of benchmark is desired, which ideally should meet the following two requirements: 

(1) \textbf{Purity}. Different from sequences in existing tracking datasets with many challenge factors, each sequence of the desired benchmark should be involved with one and only one major factor (\ie, {\it one-factor-per-sequence}). By doing so, we are able to eliminate influences from others when examining tracking algorithms on a specific challenge factor, leading to more faithful analysis. 

(2) \textbf{Quantity}. Evaluation and analysis of trackers on a benchmark are desired to be general. To this end, a dataset should be large scale to guarantee the statistical significance of analysis, so is the desired dataset. In particular, the desired benchmark should contain {\it at least} a decent number (30 is used in our work) of videos for each challenge factor for statistically meaningful analysis~\cite{lehmann2004elements}.

\subsection{Contribution}

Motivated by the above discussion, we introduce a novel toolkit, referred to as {\it TracKlinic}, aiming to offer an in-depth understanding of visual tracking algorithms on each challenge factor. TracKlinic consists of two components focusing on data and analysis aspects, respectively. 

\vspace{1mm}\noindent
{\bf Data Component.} At the core of the proposed TracKlinic is an elaborately designed benchmark, which is generated by first {\it per frame} labeling 461 videos of three benchmarks (OTB-2015~\cite{wu2015object}, TC-128~\cite{liang2015encoding} and LaSOT$_{\mathrm{tst}}$~\cite{fan2019lasot}) with seven challenge factors. In total, 776K frames are labeled with 7$\times$776K annotations. With these annotations, we construct the TracKlinic diagnosis benchmark by extracting eligible (sub-)sequences from the 461 annotated ones. Eventually, TracKlinic consists of 2,390 sequences with 280K frames. Each sequence in TracKlinic consists of {\it one and only one} major factor\footnote{In practice, there is a small chance that a sequence still suffers from some unidentified factors, but statistically negligible in our study.}, which is essentially different from existing benchmarks (see Fig.~\ref{fig1} (b)). By doing so, we can reliably examine a tracker on a specific factor by eliminating noisy influences from others, providing guidance for pertinent improvements in future research. Besides, TracKlinic allows more fair comparison of trackers on these factors.

The annotation above involves seven \textit{simple} challenge factors, including {\it Occlusion}, {\it Background Clutter}, {\it Illumination Variation}, {\it Motion Blur}, {\it Out-of-View}, {\it Rotation} and {\it Shape Variation}. We observe that, not surprisingly, some factors often accompany with each other, and may be worth being investigated together. Thus, we include into TracKlinic two \textit{compound} challenge factors, {\it Occlusion with Background Clutter} and {\it Occlusion with Rotation}, both of which are seriously under-studied in previous work. Finally, nine challenge factors are included in TracKlinic. 

\vspace{1mm}\noindent
{\bf Analysis Component.} For the analysis aspect, we present a diagnosis tool. Given the tracking results of all sequences in TracKlinic, this tool analyzes the behavior of a tracker on each individual factor and generates a report automatically. In specific, we demonstrate failure distribution of a tracker and examine its robustness to each challenge factor. In this way, we can better understand the strengths and weaknesses of a tracker, allowing of pertinent improvement. In addition, we conduct more reliable comparisons of trackers on each factor by erasing noise caused by irrelevant ones. Moreover, we measure the consistency of a tracker to different factors to validate its stability.


Integration of the above two components forms our novel diagnosis toolkit TracKlinic. We exemplify its use and capability on ten state-of-the-art trackers. Our analysis reveals that, heavy {\it shape variation} and {\it occlusion} are the two most challenging factors faced by most trackers. In addition, {\it out-of-view}, though does not occur frequently, it is often fatal to visual trackers, and effective strategies should be presented to handle it. More analysis is detailed in later sections.

In summary, our contributions are two-fold:

{\bf (\romannumeral1)} We propose the novel TracKlinic for investigating per factor tracking behaviors. Our toolkit offers the community a diagnosis dataset consisting of 2,390 sequences with 280K frames. To the best of our knowledge, TracKlinic is the first benchmark for diagnosing challenge factors in tracking. In addition, an analysis tool is presented to study trackers and generate report automatically.

{\bf (\romannumeral2)} We apply our TracKlinic to ten state-of-the-art trackers, which serves as an example of how researchers can diagnose their own trackers. We include detailed analysis for the exemplification purpose, beyond any insights observed from the diagnosis results.

The proposed diagnosis toolkit TracKlinic and analysis will be made public upon the publication of this work.



\section{Related Work}


\vspace{-1.01mm}
\paragraph{Visual Tracking Benchmarks}

Benchmarks have played a crucial role in advancing the tracking research. For a long time, tracking algorithms are usually evaluated with a small number of sequences, resulting in the problem of subjective bias~\cite{pang2013finding}. Addressing this issue, various tracking benchmarks are proposed, such as OTB~\cite{wu2013online,wu2015object}, TC-128~\cite{liang2015encoding}, VOT series~\cite{kristan2016novel}, NUS-PRO~\cite{li2015nus}, NfS~\cite{kiani2017need}, UAV123~\cite{mueller2016benchmark}, CDTB~\cite{lukezic2019cdtb} and ALOV~\cite{smeulders2013visual}. Recently, to provide enough data for training deep trackers, larger-scale datasets have been proposed, including OxUvA~\cite{valmadre2018long}, GOT-10K~\cite{huang2018got}, LaSOT~\cite{fan2019lasot} and TrackingNet~\cite{muller2018trackingnet}. These benchmarks greatly advance the research of visual tracking. Nevertheless, as discussed above, they are not suitable for reliable diagnosis of tracking algorithms regarding specific challenge factors. 

Different from these benchmarks, the proposed TracKlinic is dedicated for challenge factor diagnosis. For this goal, \textit{per frame} challenge factor annotation is first provided on initial sequences. Moreover, one-factor-per-sequence is enforced in TracKlinic for reliable analysis. These characteristics make TracKlinic uniquely suitable for the diagnosis purpose, and enable us, for the first time (to our knowledge), to perform per factor analysis. 


It is worth noting that, TracKlinic does not aim to replace existing tracking benchmarks. Instead, it can be viewed as complimentary to existing datasets to focus on challenge factor analysis.

\vspace{-2.51mm}
\paragraph{Diagnosis of Tracking Algorithms}

Another important related work is~\cite{wang2015understanding}. In \cite{wang2015understanding}, a diagnosis approach is proposed to analyze the impact of each tracking component on final tracking performance. Specifically, five constitution parts including motion model, feature extractor, observation model, model updater and ensemble post-processor are examined using different implementations, aiming to offer guidance to researchers in designing robust trackers.

Our work is significantly different from~\cite{wang2015understanding}. The work of~\cite{wang2015understanding} focuses on studying the impact of a specific tracking component on tracking performance, while ours examines the abilities of a tracker on different challenge factors. In fact, the diagnosis approach of~\cite{wang2015understanding} and this work are complementary and can work together to provide directions for improving tracking performance.

\vspace{-2.51mm}
\paragraph{Diagnosis in Other Vision Problems}

Our work is also closely related to similar studies~\cite{hoiem2012diagnosing,zhang2016far,ruggero2017benchmarking,sigurdsson2017actions,alwassel2018diagnosing,XingLHYL17pr} for other tasks. The seminal work of~\cite{hoiem2012diagnosing} explores different failure modes in object detection and showcases the impacts of different characters on detection performance. The approach of~\cite{zhang2016far} analyzes localization errors for pedestrian detection. The methodology of~\cite{ruggero2017benchmarking} provides an insightful diagnosis of different algorithms in the field of action understanding in videos and discusses relevant directions for future improvements. The method of~\cite{sigurdsson2017actions} introduces the diagnosis analysis into human pose estimation with the goal of quantitatively identifying the failure modes of different algorithms and recommending effective strategies for improvements in body part localization. In~\cite{alwassel2018diagnosing}, a diagnostic tool is presented to characterize errors for temporal action localization. The work of~\cite{XingLHYL17pr} diagnoses deep learning models for age estimation.

In a similar spirit with these studies, we present a diagnosis approach to characterize different challenge factors in tracking, and thus help researchers to understand the performance of a visual tracking algorithm on specific challenge factors.


\section{TracKlinic}

In this section, we will detail the proposed toolkit TracKlinic, including a diagnosis benchmark and an analysis tool.

\subsection{Diagnosis Dataset}

In order to validate effectiveness of a tracker in handling a specific challenge factor, it is essential to inhibit disturbance from others in videos. However, the sequences in existing datasets are often involved with more than one challenge factor (see Fig.~\ref{fig1} (a)). In such situation, it is not clear which factor is the cause of a tracking failure. To address this issue, we propose TracKlinic, which, to the best of our knowledge, is the first diagnosis dataset in the tracking community. 

\vspace{-0.9em}
\subsubsection{Initial Collection}
\vspace{-0.5em}

Our goal is {\it not} to build a completely new dataset for assessing overall performance of a tracker as in existing datasets. Instead, we aim to diagnose an algorithm on different challenge factors to further improve its performance on existing benchmarks. Considering this, it is natural to adopt existing benchmarks as our data source.

\begin{figure}[!t]
	\centering
	\includegraphics[width=0.95\linewidth]{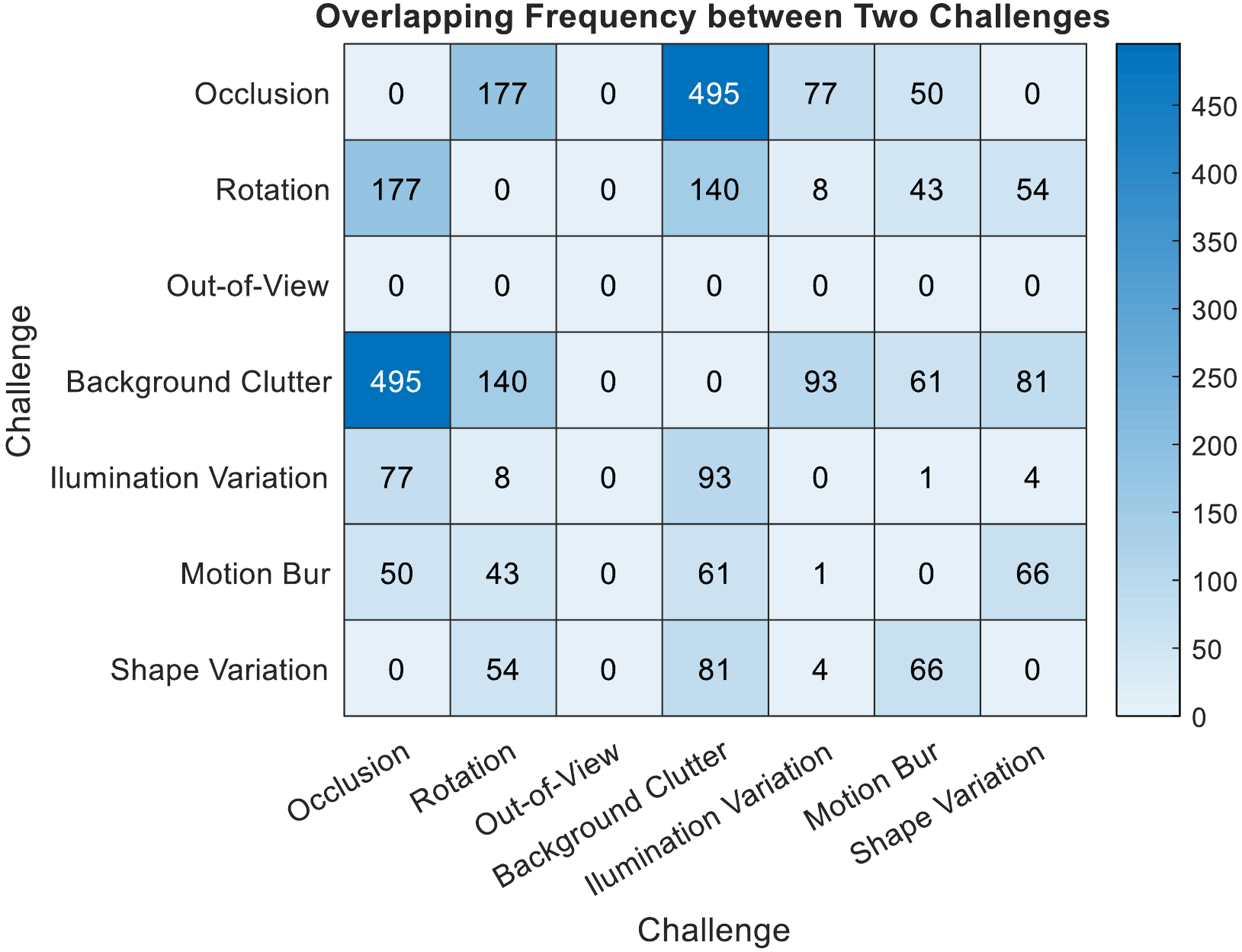}\\
	\caption{The statistic of overlapping frequency between challenge factors in OTL. Best viewed in color.}
	\label{fig2}
\end{figure}

Since this work is focused on generic model-free single-object tracking, 
we narrow down our choices to OTB-2015~\cite{wu2015object}, TC-128~\cite{liang2015encoding}, NUS-PRO~\cite{li2015nus}, VOT series~\cite{kristan2016novel}, GOT-10K~\cite{huang2018got}, LaSOT~\cite{fan2019lasot} and TrackingNet~\cite{muller2018trackingnet}. With a joint consideration of target diversity, groundtruth availability and the amount of required annotation, we choose OTB-2015~\cite{wu2015object}, TC-128~\cite{liang2015encoding} and LaSOT$_\mathrm{tst}$~\cite{fan2019lasot} as our data source. 
For conciseness, we use OTL to denote these three benchmarks. In this way, we obtain 461 videos with 770K frames in OTL for initial collection of TracKlinic.

\vspace{-0.9em}
\subsubsection{Per-frame Factor Annotation}
\vspace{-0.5em}

Tracking is a complex process, and it is difficult to enumerate all challenge factors that may result in tracking failures. In this paper, we study the seven most common factors in tracking, including {\it Occlusion} (OCC), {\it Background Clutter} (BC), {\it Illumination Variation} (IV), {\it Motion Blur} (MB), {\it Out-of-View} (OV), {\it Rotation} (ROT) and {\it Shape Variation} (SV)\footnote{It is worth noting that, since the {\it Deformation} and {\it Aspect Ratio Change} almost always accompany with {\it Scale Variation}, we unify all these three challenge factors into {\it Shape Variation}.}.

Unlike conventional tracking benchmarks in which each video is labeled with global challenge factors, our goal is to build a new type of dataset where each sequence is involved with {\it one and only one} major factor. For this purpose, we need to label each video in OTL with {\it per frame} challenge factors. Based on these annotations, we extract eligible subsequences from OTL and ensure that each one qualifies the {\it one-factor-per-sequence} rule. 

\renewcommand\arraystretch{1.1}
\begin{table}[!t]\scriptsize
	\centering
	\caption{Descriptions of nine challenge factors in OTL.}
	\begin{tabular}{p{10em}p{19em}}
		\hline
		Challenge Factor & Definition  \\
		\hline \hline
		{Occlusion (OCC) }   & Target is occluded in a frame    \\
		{Rotation (ROT)}  & Target rotates in a frame   \\
		{Out-of-View (OV)}   & Target completely leaves the video frame\\
		{Bkg. Clut. (BC)}   & The background is similar to target\\
		{Illumination Vari. (IV) }   & The illumination in target region changes \\
		{Motion Blur (MB)}   & Target region is blurred due to motion\\
		{Shape Variation (SV)}   & The ratio of bounding box or its aspect ratio is outside the range [0.25, 4]\\
		{Occ.-Bkg. Clut. (O-B)}   & Target suffers from overlapping OCC and BC\\
		{Occ.-Rot. (O-R)}   & Target suffers from overlapping OCC and ROT\\
		\hline
	\end{tabular}%
	\label{tab:definition}%
\end{table}%

To guarantee the annotation quality, each frame is manually labeled with the aforementioned challenge factors (except for SV) by an expert and then verified by another expert. The annotation of SV is automatically computed based on its definition. It is worth noticing that, since we focus on diagnosing challenges in object tracking, a frame is labeled with challenge factors only when obvious target appearance variations are caused by them. In total, we finish 7$\times$776K manual annotations for all sequences in OTL. An example of our annotation for a video is shown in Fig.~\ref{fig1} (a) and more video examples can be seen in {\bf supplementary material}.

In OTL, each challenge factor represents a type of challenge (referred to as {\it simple-challenge} factor) for tracking. However, it is rather frequent in tracking that two or more factors occur simultaneously. Considering this, we also study challenge factors containing two types of challenges (referred to as {\it compound-challenge} factor). In specific, we count the overlapping frequency between two challenges in OTL. Note that, in order to be considered as overlapping, the number of overlapping frames of two challenges has to be greater than $\tau_{\mathrm{op}}$ (set to 3). Fig.~\ref{fig2} demonstrates the statistical results. In this work, we select the two most frequent compound-challenge factors, \ie, {\it Occlusion with Background clutter} (O-B) and {\it Occlusion with Rotation} (O-R). After that, we automatically generate annotations for these two compound-challenge factors and meanwhile adjust annotations for OCC, BC and ROT. 

In summary, we diagnose nine factors for tracking algorithms. The definitions of these factors are shown in Tab.~\ref{tab:definition}.

\vspace{-0.9em}
\subsubsection{Single-factor Sequence Extraction}
\vspace{-0.5em}

With the per-frame annotations of each challenge factor, we extract single-factor sequence from OTL under {\it one-factor-per-sequence} rule. These eligible videos form our diagnosis dataset. We describe the extraction rules as follows.

\begin{figure*}[!t]\small
	\centering
	\includegraphics[width=1\linewidth]{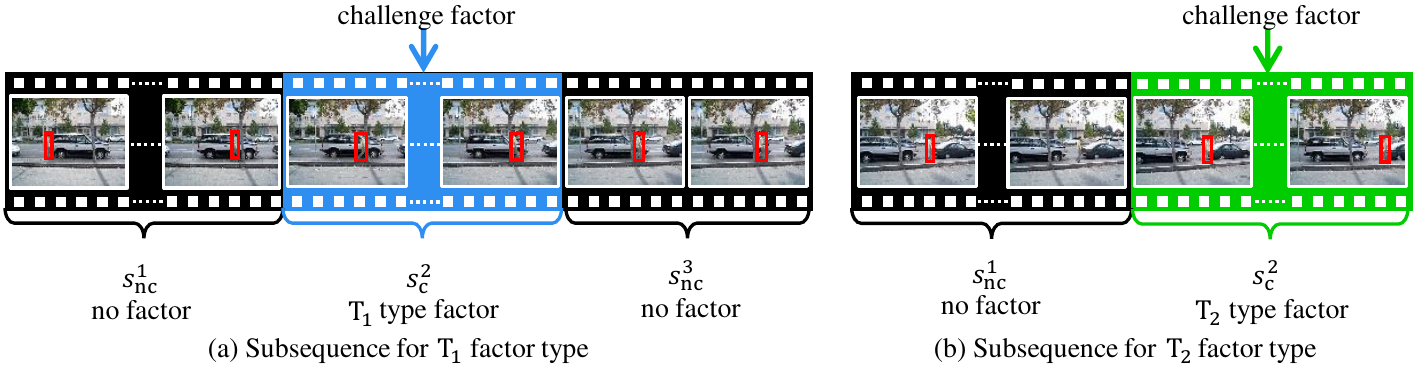}\\
	\caption{Generating subsequences of TracKlinic for challenge factors of types $\mathrm{T}_{1}$ and $\mathrm{T}_{2}$. Best viewed in color and by zooming in.}
	\label{fig3}
\end{figure*}

For extraction purpose, we classify challenge factors into two types $\mathrm{T}_{1}$ and $\mathrm{T}_{2}$ based on the status of last frame in a challenge. If the last frame suffers from occlusion or out-of-view, the challenge factor belongs to $\mathrm{T}_{1}$, otherwise $\mathrm{T}_{2}$. For challenge factor of $\mathrm{T}_{1}$, we extract subsequences which satisfy: the subsequence can be decomposed into three consecutive parts $s_{\mathrm{nc}}^{1}$, $s_{\mathrm{c}}^{2}$ and $s_{\mathrm{nc}}^{3}$, where $s_{\mathrm{nc}}^{1}$ and $s_{\mathrm{nc}}^{3}$ do not contain any challenge factors and $s_{\mathrm{c}}^{2}$ exclusively contains this factor, and the lengths $|s_{\mathrm{nc}}^{1}| \geq \tau_{\mathrm{s}}$ (set to 10) and $|s_{\mathrm{nc}}^{3}| = \tau_{\mathrm{e}}$ (set to 2), as shown in Fig.~\ref{fig3} (a). Considering that long sequence may cause more cumulative errors, we only retain the last 30 frames in $s_{\mathrm{nc}}^{1}$ if $|s_{\mathrm{nc}}^{1}|>30$ for all factors except for shape variation, because 30 frames are sufficient for stable tracker initialization. For challenge factor of $\mathrm{T}_{2}$, we extract subsequences that satisfy: the subsequence can be divided into two consecutive parts $s_{\mathrm{nc}}^{1}$ and $s_{\mathrm{c}}^{2}$, where $s_{\mathrm{nc}}^{1}$ does not contain any challenge factors and $s_{\mathrm{c}}^{2}$ exclusively contains this factor, and the length $|s_{\mathrm{nc}}^{1}| \geq \tau_{\mathrm{s}}$, as shown in Fig.~\ref{fig3} (b). Likewise, we only retain the last 30 frames in $s_{\mathrm{nc}}^{1}$ if $|s_{\mathrm{nc}}^{1}|>30$. The difference between $\mathrm{T}_{1}$ and $\mathrm{T}_{2}$ is that, for challenge factor of $\mathrm{T}_{1}$, the target may be partly visible (\eg, occlusion) or completely invisible (\eg, out-of-view) in the last frame. In such case, it is not suitable to estimate target status. Therefore, we add an extra part $s_{\mathrm{nc}}^{3}$ to $\mathrm{T}_{1}$ type factor to ensure that the target is fully visible. 

\begin{figure}[!t]
	\centering
	\includegraphics[width=0.95\linewidth]{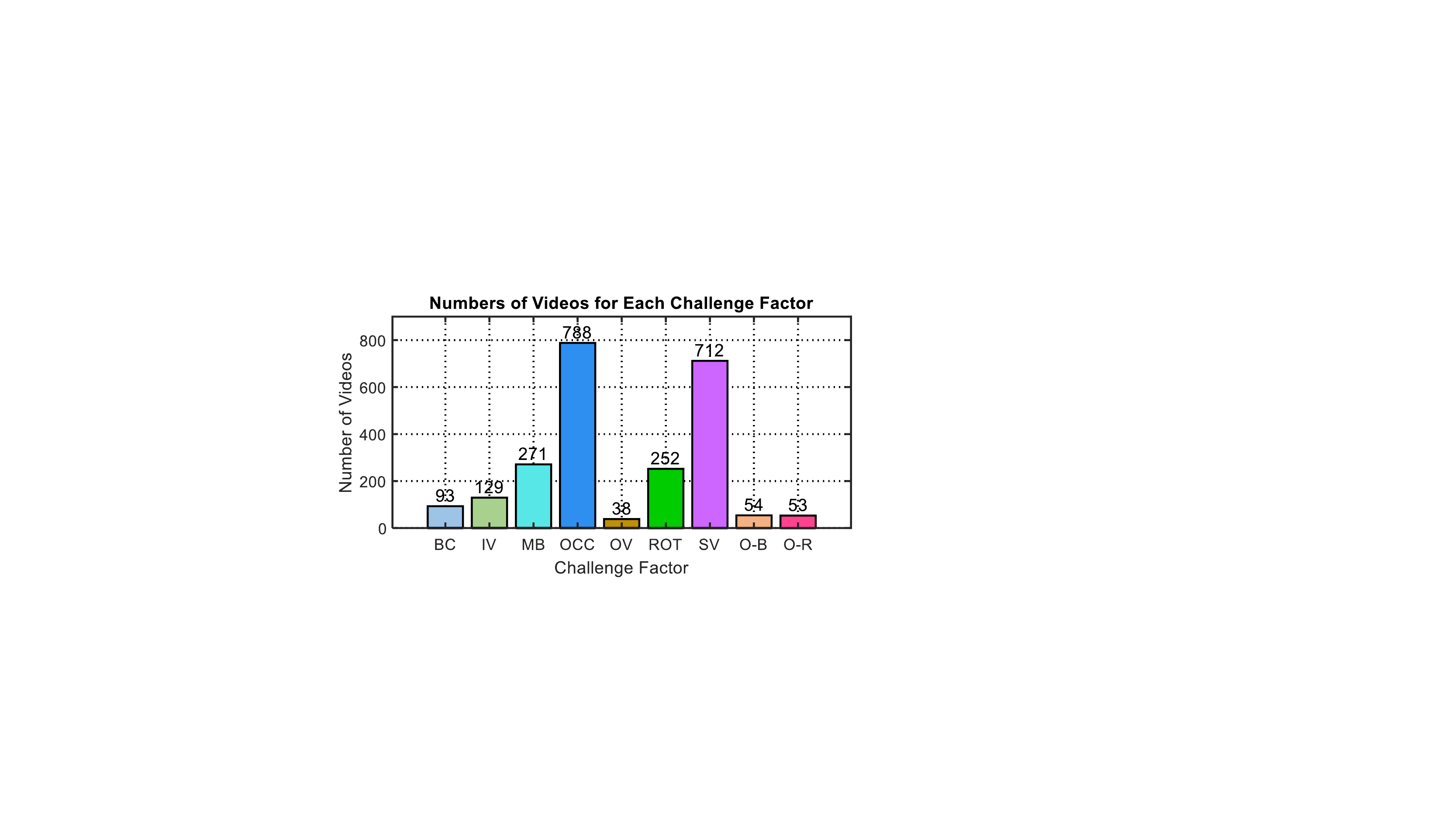}\\
	\caption{Numbers of sequences for different challenge factors in TracKlinic. There are {\it at least} 30 videos for each challenge factor, indicating that our diagnosis analysis is statistically meaningful. }
	\label{fig4}
\end{figure}

\renewcommand\arraystretch{1.15}
\begin{table}[!t]\scriptsize
	\centering
	\caption{Comparison of TKB and other tracking Benchmark. OTL combines OTB-2015~\cite{wu2015object}, TC-128~\cite{liang2015encoding} and LaSOT$_{\mathrm{tst}}$~\cite{fan2019lasot}.}
	\begin{tabular}{@{}R{2cm}||@{}C{0.8cm}@{}C{0.9cm}@{}C{0.9cm}@{}C{0.9cm}@{}C{0.9cm}@{}C{1.4cm}@{}}
		\hline
		Benchmark & Videos & \tabincell{c}{Min \\ frames} & \tabincell{c}{Mean \\ frames} & \tabincell{c}{Max \\ frames} & \tabincell{c}{Total \\ frames} & \tabincell{c}{Chal. \\ factor anno.} \\
		\hline
		\hline
		VOT-18~\cite{kristan2018sixth} & 60   & 41    & 356   & 1,500  & 21$\mathbf{K}$   & n/a \\
		UAV123~\cite{mueller2016benchmark} & 123  & 109    & 915   & 3,085  & 113$\mathbf{K}$   & global \\
		NfS~\cite{kiani2017need} & 100  & 169    & 3,830   & 20,665  & 383$\mathbf{K}$   & global \\
		GOT-10K$_{\mathrm{tst}}$~\cite{huang2018got} & 180   & 51    & 127   & 920  & 23$\mathbf{K}$   & n/a \\
		TrackingNet$_{\mathrm{tst}}$~\cite{muller2018trackingnet} & 511   & 96    & 441   & 2,368  & 226$\mathbf{K}$   & n/a \\
		OTB-2015~\cite{wu2015object} & 100   & 71    & 590   & 3,872  & 59$\mathbf{K}$   & global \\
		TC-128~\cite{liang2015encoding} & 129   & 71    & 429   & 3,872  & 55$\mathbf{K}$   & global \\
		LaSOT$_{\mathrm{tst}}$~\cite{fan2019lasot} & 280   & 1,000  & 2448  & 9,999  & 690$\mathbf{K}$  & global \\
		OTL   & 461   & 71    & 1683  & 9,999  & 776$\mathbf{K}$  & per frame \\
		\hline
		\hline
		TracKlinic & 2,390  & 11    & 115   & 6,559  & 280$\mathbf{K}$  & per frame \\
		\hline
	\end{tabular}%
	\label{tabl}%
\end{table}%
After the above procedure, we construct TracKlinic with each sequence involving {\it one and only one} major factor. We show examples for each factor in {\bf supplementary material}. Tab.~\ref{tabl} summarizes TracKlinic and comparisons with other datasets. Fig.~\ref{fig4} shows the numbers of videos for each factor.

\subsection{Analysis Tool}

In addition to the benchmark, an analysis methodology is presented in our toolkit for studying trackers on each factor. 

In specific, we demonstrate the proportion of failures due to challenge factors for trackers, which allows us to understand the potential failure causes. In addition, we study the failure rates of trackers for each factor. In this way, we can better understand the strengths and weaknesses of trackers for future improvements. Furthermore, we compare different tracking algorithms on our benchmark. In comparison with other datasets, our analysis is more reliable because noise caused by irrelevant factors are erased. We also measure consistency of a tracker to different factors.

Given the tracking results of all sequences in TracKlinic, our tool analyzes the above per factor tracking behavior and generates the report automatically, with which researchers can further improve their own trackers.


\section{Diagnosis of Challenge Factors}

\subsection{Algorithms}

\begin{figure*}[!t]
	\centering
	\includegraphics[width=\linewidth]{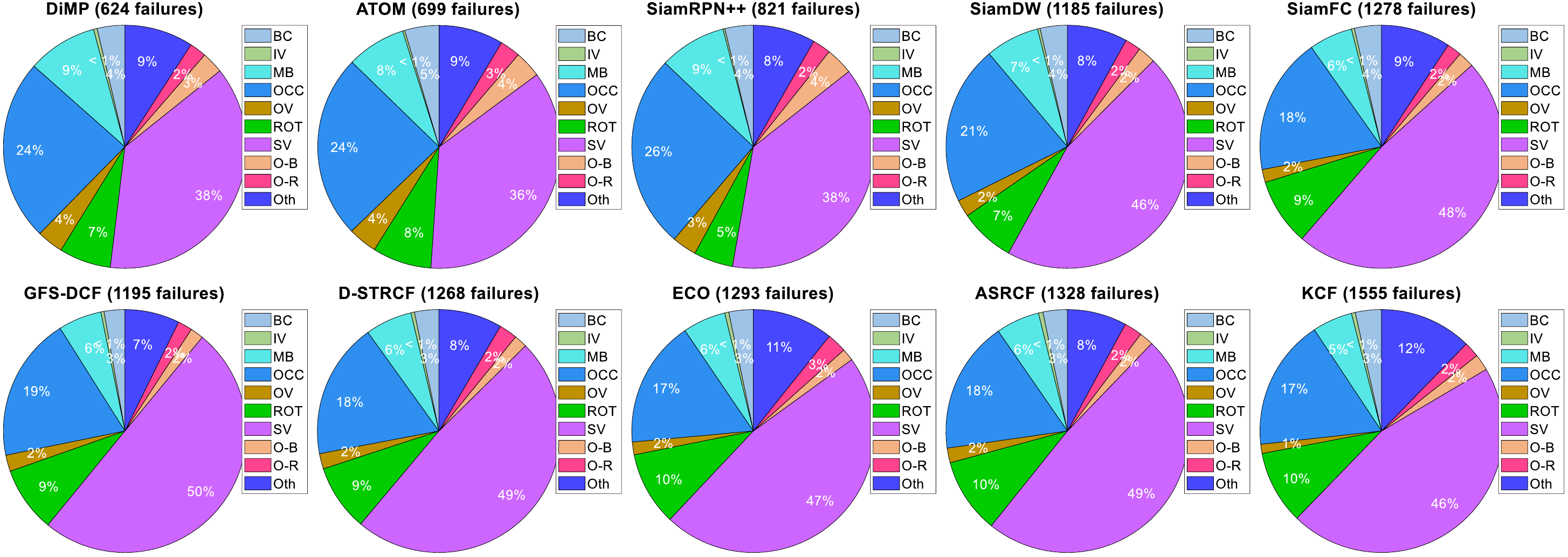}\\
	\caption{The percents of failures caused by different challenge factors for ten trackers. Best viewed in color and by zooming in.}
	\label{fig5}
\end{figure*}

We apply TracKlinic to ten state-of-the-art trackers. In specific, we choose the top three trackers with available implementations on each of the four datasets, including VOT-19~\cite{kristan2019seventh}, LaSOT~\cite{fan2019lasot}, OTB-2015~\cite{wu2015object} and TC-128~\cite{liang2015encoding}. We exclude the repeated ones from these trackers. In this way, we obtain eight algorithms, including DiMP~\cite{bhat2019learning}, ATOM~\cite{danelljan2019atom}, SiamRPN++~\cite{li2019siamrpn++}, SiamDW~\cite{zhang2019deeper}, ASRCF~\cite{dai2019visual}, ECO~\cite{danelljan2017eco}, D-STRCF~\cite{li2018learning} and GFS-DCF~\cite{xu2019joint}. In addition, we also analyze two baseline trackers KCF~\cite{henriques2014high} and SiamFC~\cite{bertinetto2016fully} as these two trackers have led the recent trends of tracking with many extensions~\cite{ma2015hierarchical,sun2019roi,fan2017parallel,li2018high,valmadre2017end,he2018twofold,zhang2018structured,fan2019siamese}, and it is worth diagnosing them on different challenge factors.

\subsection{Proportion of Failures by Challenge Factors}

It is important to understand failures of a tracker caused by different challenge factors for improvements. In this section, we take ten trackers as an example of showcasing the proportions of failures by each factor. 

Since each video in TracKlinic is involved with one major challenge factor, we define tracking failure on a video according to the tracking result in the last frame. If the intersection over union (IoU) of tracking result with groundtruth is less than a threshold $\tau_{\mathrm{IoU}}$ (set to 0.5), the tracker fails for the challenge factor in this video. Note that, although each video contains one major factor, it is possible that a tracker fails before the challenge in a video happens. For this case, we identify the tracking failure caused by others.

Fig.~\ref{fig5} shows the proportions of failures by different challenge factor. Surprisingly, we observe that for all ten trackers, although shape variation does not happen the most frequently in videos (see Fig.~\ref{fig4}), it is the most likely to cause a failure. Specifically, the probabilities of a failure caused by the shape variation are 38\%, 36\%, 38\%, 46\%, 48\%, 50\%, 49\%, 47\%, 49\% and 46\% for DiMP, ATOM, SiamRPN++, SiamDW, SiamFC, GFS-DCF, D-STRCF, ECO, ASRCF and KCF, respectively. 

Following shape variation, occlusion is second most likely to cause a tracking failure, with chances of 24\%, 24\%, 26\%, 21\%, 18\%, 19\%, 18\%, 17\%, 18\% and 17\% for the ten trackers. Note that, shape variation and occlusion account for more than 60\% of failures, suggesting that strategies to better handle them may be the most rewarding for future research. Moreover, we can see from Fig.~\ref{fig5} that, illumination variation is the least possible reason for a failure. For all trackers, the chances of a failure caused by illumination variation do not exceed 1\%, indicating that future research can focus more on dealing with other factors such as shape variation, occlusion, rotation, motion blur and so forth.

\subsection{Understanding Trackers on Different Factors}

To understand the strengths and weaknesses of a tracker, it helps (1) guide future researches for improvement and (2) deploy the tracker in practical applications as we can choose the tracking algorithms according to the specific scenarios. In this section, we show the abilities of ten representative trackers in response to each challenge factor. We use failure rate, defined as the ratio of the number of tracking failures and the number of videos of a challenge factor, to represent the ability of the tracker. The lower the failure rate is, the stronger the tracker is in handling the challenge factor.

\begin{figure*}[!t]
	\centering
	\includegraphics[width=\linewidth]{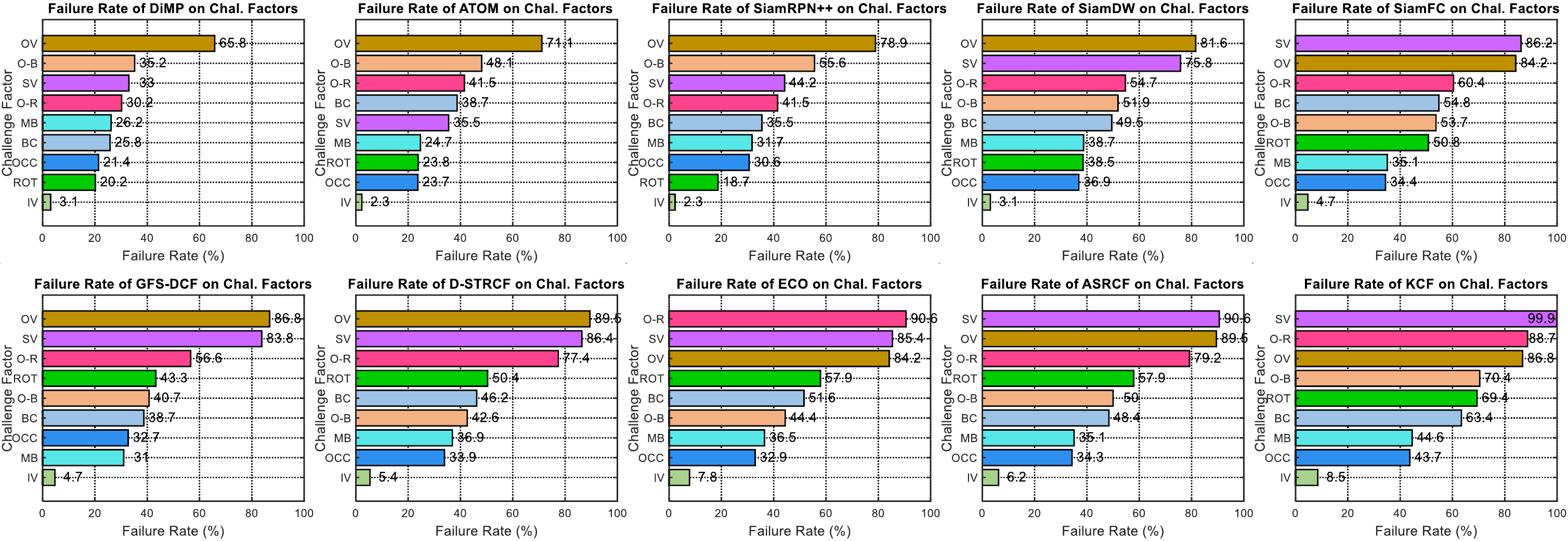}\\
	\caption{Failure rates of ten state-of-the-art trackers for each challenge factor. Best viewed in color and by zooming in.}
	\label{fig6}
\end{figure*}

\begin{figure*}[!t]
	\centering
	\begin{tabular}{@{}C{2.8cm}@{}@{}C{2.8cm}@{}@{}C{2.8cm}@{}@{}C{2.8cm}@{}@{}C{2.8cm}@{}@{}C{2.8cm}@{}}
		\includegraphics[width=2.7cm]{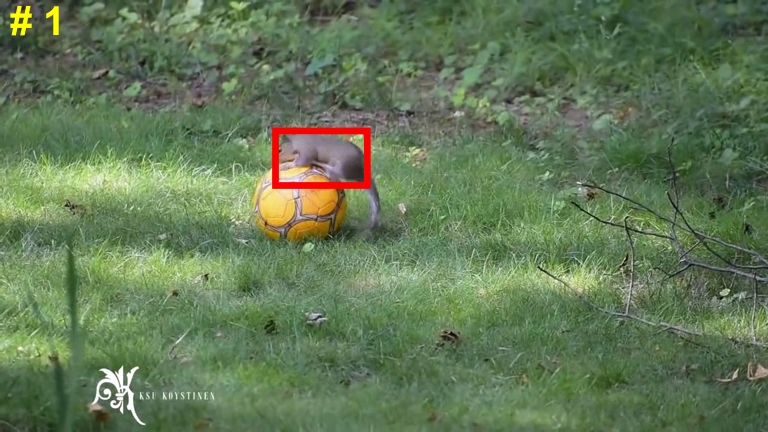} & \includegraphics[width=2.7cm]{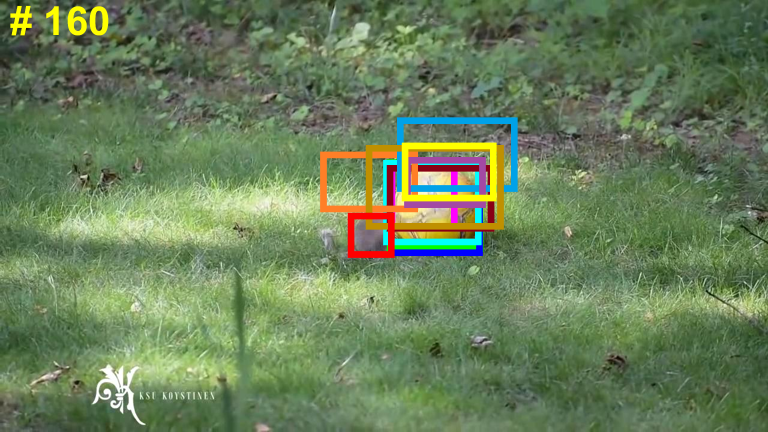} & \includegraphics[width=2.7cm]{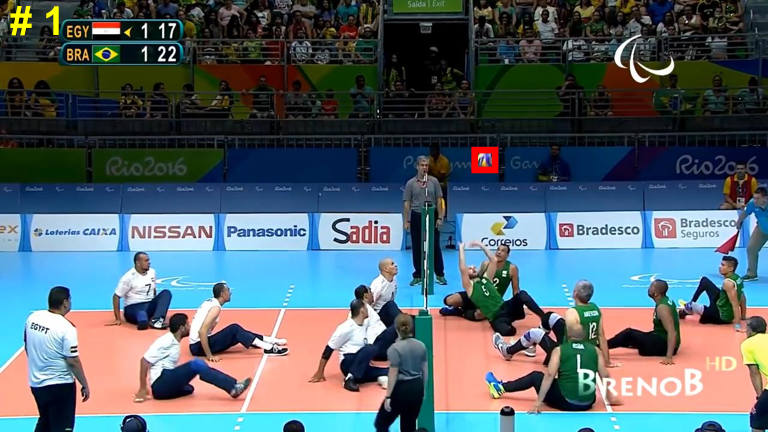} & \includegraphics[width=2.7cm]{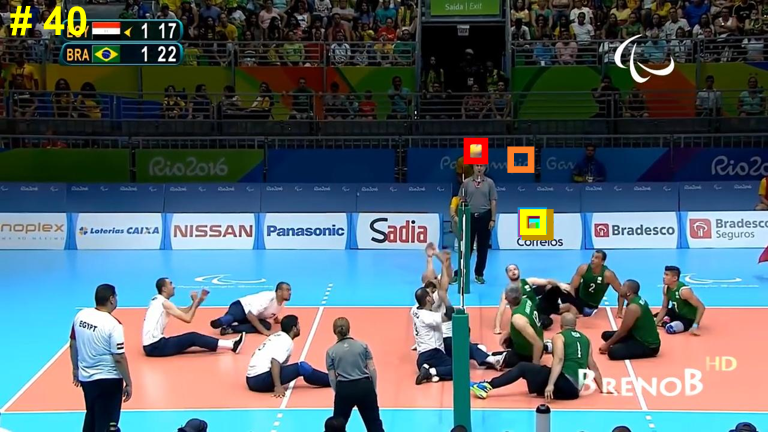} & \includegraphics[width=2.7cm]{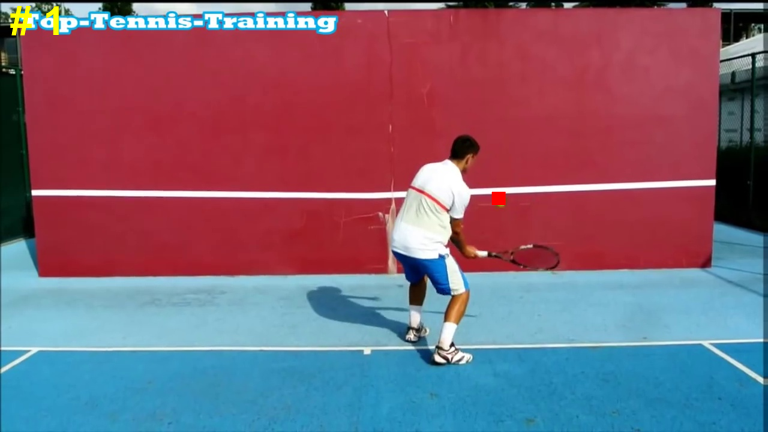}& \includegraphics[width=2.7cm]{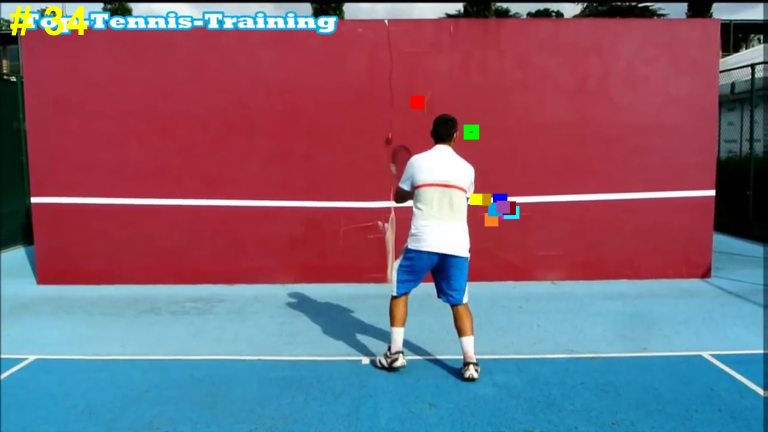} \\
		
		\multicolumn{2}{c}{(a) Video with challenge factor O-R}  & \multicolumn{2}{c}{(b) Video with challenge factor MB}  & \multicolumn{2}{c}{(c) Video with challenge factor OCC}\\

		\includegraphics[width=2.7cm]{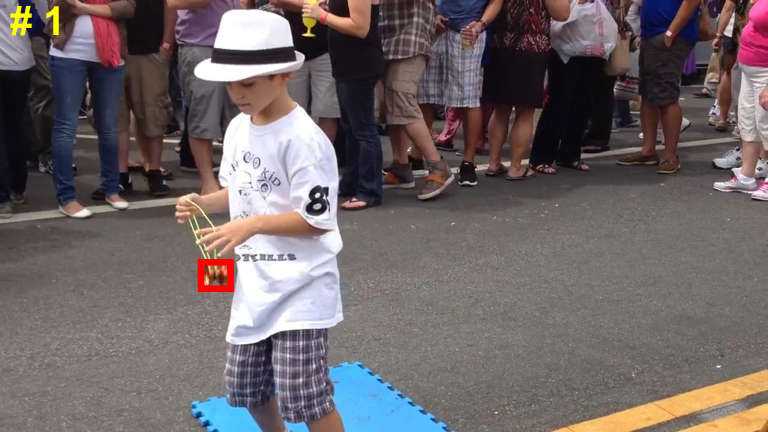} & \includegraphics[width=2.7cm]{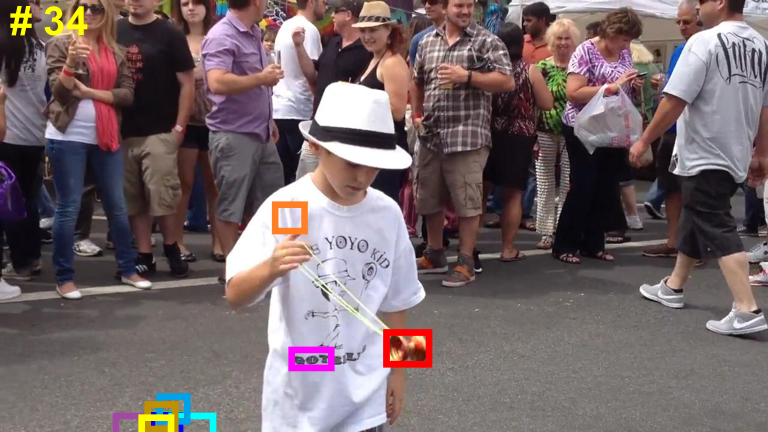} & \includegraphics[width=2.7cm]{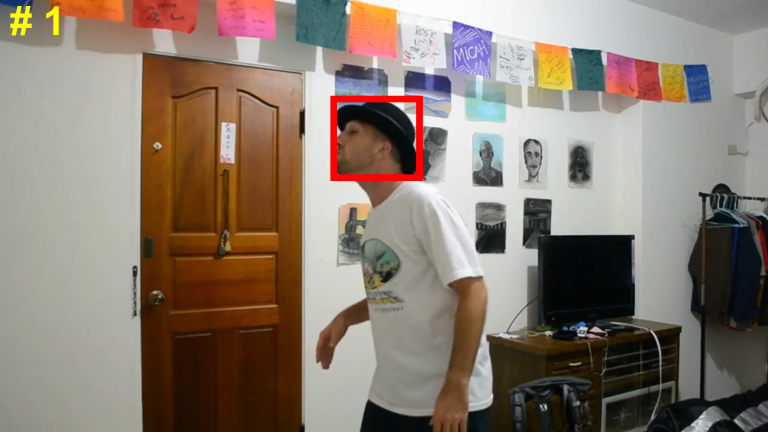} & \includegraphics[width=2.7cm]{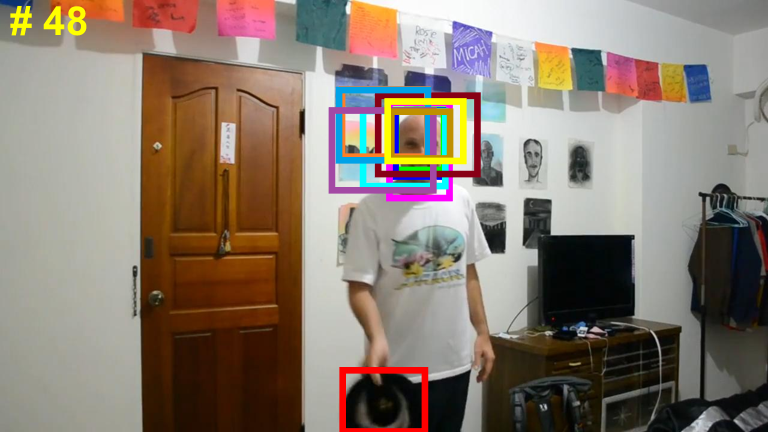} & \includegraphics[width=2.7cm]{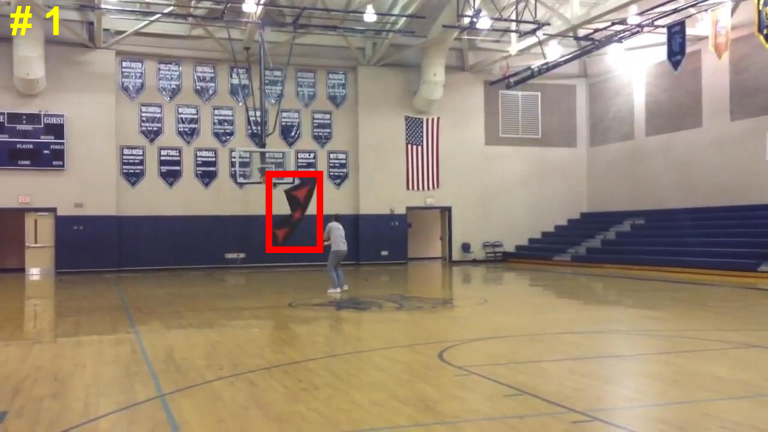}& \includegraphics[width=2.7cm]{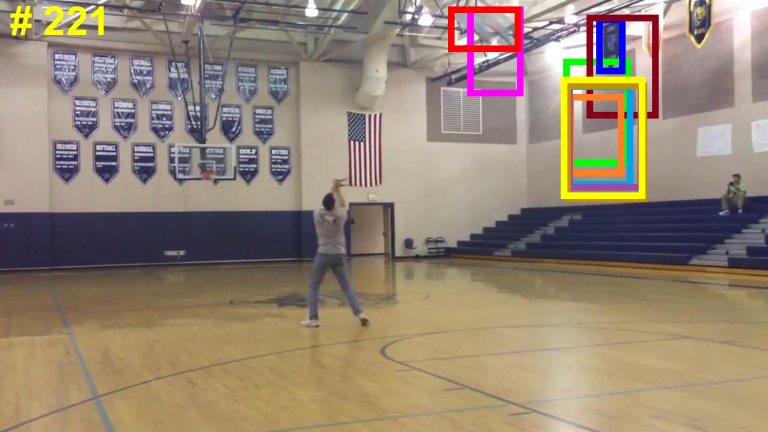} \\
		
		\multicolumn{2}{c}{(d) Video with challenge factor OV}  & \multicolumn{2}{c}{(e) Video with challenge factor ROT}  & \multicolumn{2}{c}{(f) Video with challenge factor SV}\\
		
		\includegraphics[width=2.7cm]{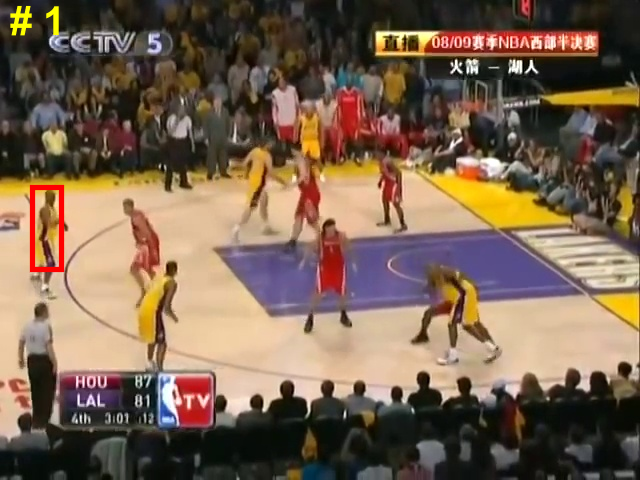} & \includegraphics[width=2.7cm]{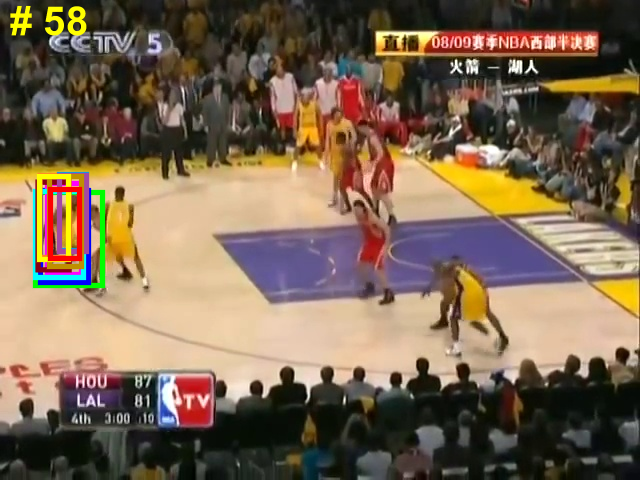} & \includegraphics[width=2.7cm]{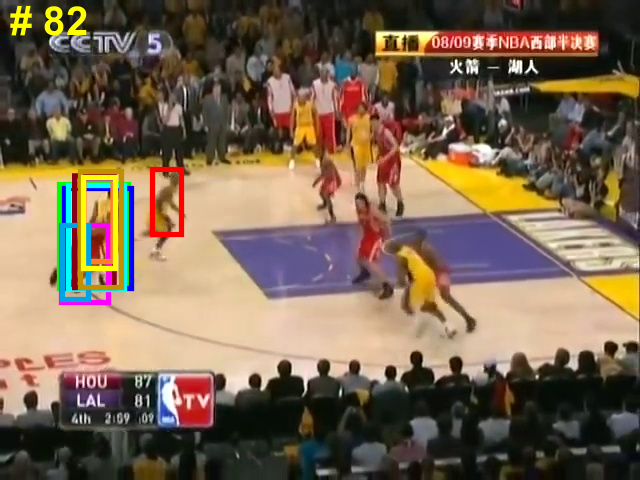} & \includegraphics[width=2.7cm]{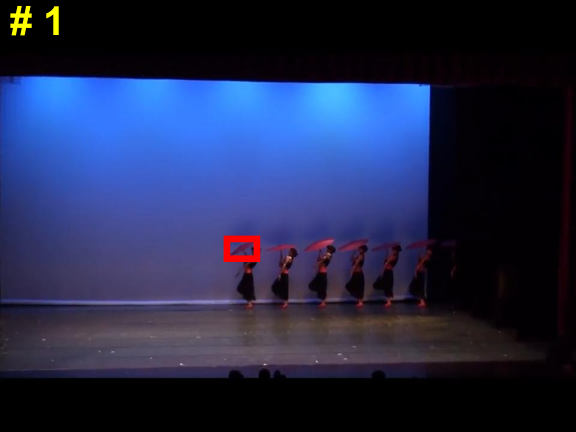} & \includegraphics[width=2.7cm]{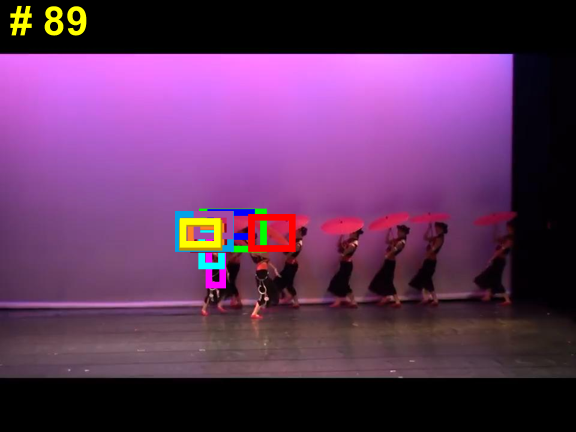} & \includegraphics[width=2.7cm]{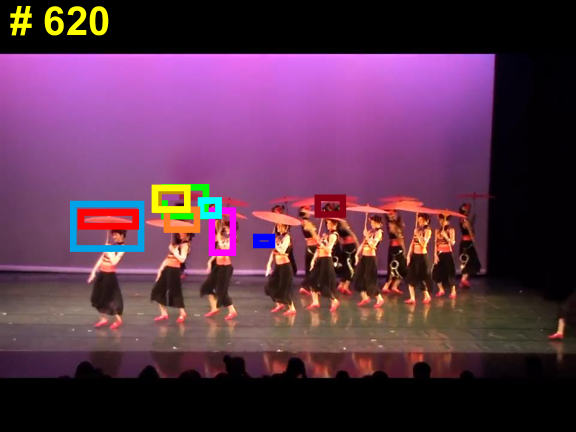} \\
		
		\multicolumn{3}{c}{(g) Video with challenge factor O-B}  & \multicolumn{3}{c}{(h) Video with challenge factor BC}  \\
		
		\multicolumn{6}{c}{\includegraphics[width=16cm]{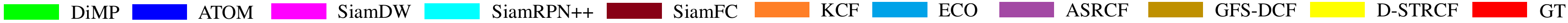}}\\
	\end{tabular}
	\caption{Qualitative results of failure cases for each tracker under eight challenge factors. Best viewed in color.}
	\label{fig7}
\end{figure*}

Fig.~\ref{fig6} shows the failure rates of ten trackers on each factor. An interesting observation is that, although out-of-view appears least frequently (see Fig.~\ref{fig4}), it is fatal to trackers. For DiMP, ATOM, SiamRPN++, SiamDW, SiamFC, GFS-DCF, D-STRCF, ECO, ASRCF and KCF, their failures rates under the out-of-view are 65.8\%, 71.1\%, 78.9\%, 81.6\%, 86.2\%, 86.8\%, 89.5\%, 84.2\%, 89.5\% and 86.8\%, respectively. It is worth noting that DiMP, ATOM, SiamRPN++ and SiamDW utilize very deep networks~\cite{he2016deep} for feature representation. Nevertheless, they still easily fail when out-of-view happens, which implies the requirement of a special mechanism in handling out-of-view for tracking. A feasible solution is to couple the tracker with a global detector~\cite{kalal2011tracking,yan2019skimming} which can re-locate the target when it appears again in the view. 

In addition, shape variation is difficult to visual trackers, especially for the baseline SiamFC and correlation filter based ones including GFS-DCF, D-STRCF, ECO, ASRCF and KCF. The failure rates of these trackers are higher than 80\% due to the lack of an effective strategy to estimate target state. This issue is alleviated by borrowing techniques such as RPN~\cite{ren2015faster} (\ie, SiamRPN++ and SiamDW) and IoU-Net~\cite{jiang2018acquisition} (\ie, DiMP and ATOM) from object detection. Despite this, a tracker is prone to fail when shape variation occurs. Especially for non-rigid targets, the bounding-box based trackers may introduce a large amount of background information into the tracking model, resulting in gradual degradation. To achieve accurate target state estimation, a better way is to leverage the mask of target through segmentation~\cite{wang2019fast}. Although the chance of an occlusion-causing failure is high, trackers perform surprisingly robustly to this challenge factor. In particular, using deeper feature can further reduce the failure rate. Nevertheless, when accompanied with other factors such as rotation or background clutter, occlusion becomes more challenging. Furthermore, we also see that all visual trackers exhibit robust performance with failure rates of 3.1\%, 2.3\%, 2.3\%, 3.1\%, 4.7\%, 4.7\%, 5.4\%, 7.8\%, 6.4 and 8.5\% in tackling illumination variation, which indicates that the illumination variation issue is almost addressed using deep features.

In order to qualitatively analyze each tracker, we demonstrate qualitative results of common failure cases for each tracker on eight difficult factors in Fig.~\ref{fig7}.

\subsection{Challenge Factor-based Comparison}

In addition to diagnosis of each challenge factor in tracking, our TracKlinic allows more reliable comparison of different approaches for each challenge factor. We adopt success score (at the threshold of 0.5)~\cite{wu2013online} as evaluation metric.

\begin{figure*}[!tb]
	\centering
	\includegraphics[width=\linewidth]{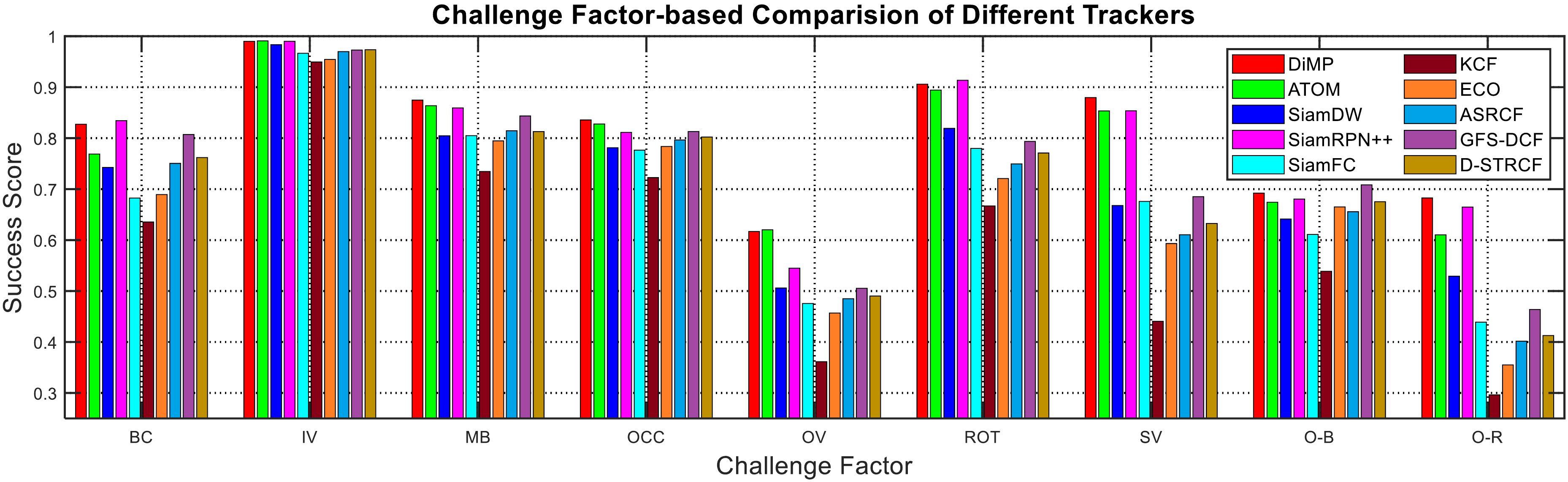}\\
	\caption{Challenge Factor-based comparison of ten state-of-the-art trackers. Best viewed in color.}
	\label{fig8}
\end{figure*}

Fig.~\ref{fig8} demonstrates the challenge factor-based comparison of different algorithms. We observe that DiMP achieves the best performance under eight out of nine challenge factors. In particular, compared with its baseline ATOM, the improvement under background clutter is obvious, which evidences the effectiveness of exploring background information in distinguishing target from distractors. For shape variation, the three trackers DiMP, ATOM and SiamRPN++ significantly outperform others, showing the importance of deeper features and the advantages of bounding box regression and IoU prediction in target state estimation. Besides, an interesting finding is that, as the two most popular approaches for state estimation, our analysis shows that IoU-Net based strategy performs slightly better than RPN based method. For illumination variation, all trackers show robust performance, which is consistent with previous diagnosis.

\begin{figure*}[!t]
	\centering
	\includegraphics[width=\linewidth]{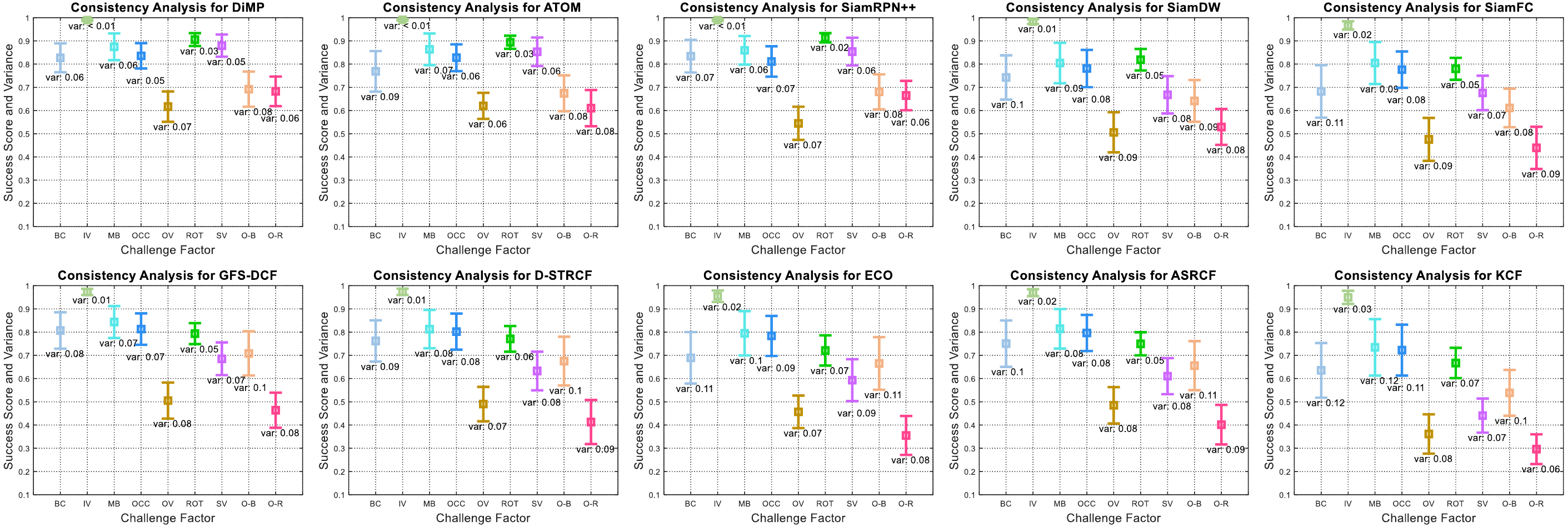}\\
	\caption{Consistency measurement of each algorithm on different challenge factors. The square represents the average success score of a tracker. The difference between the top (or bottom) and the square denotes the variance under the challenge factor. Best viewed in color.}
	\label{fig9}
\end{figure*}

\subsection{Consistency Analysis}

In real-world sceneries, it is difficult to estimate the complexities of challenge factors in a video. Therefore, a practical algorithm should demonstrate consistent (good) performance no matter how complicated the factor is. For this purpose, we conduct consistency analysis for each tracker. In detail, we adopt the variance of all success scores for consistency measurement of a tracker to a challenge factor. The smaller the variance is, the more consistent the performance is. It is worth noticing that, the consistency of a tracker may be high if it performs poorly on all challenge factors. Thus, consistency analysis should be applied with other metrics such as precision and success scores.

Fig.~\ref{fig9} demonstrates the consistency of each tracker on different factors. We observe that deep trackers performs consistently on illumination variation. However, the performance of trackers on out-of-view and background clutter is inconsistent due to their extreme complexities. For instance for out-of-view, the target may re-enter the view from a distant position. In such case, visual trackers easily fail. In a video with complicated background clutter, there may exist many distractors with identical appearances to the target. As a consequence, trackers may drift to other similar distractors. In summary, more efforts are needed to improve the consistency to challenge factors to make visual trackers practical.    


\section{Conclusion}

In this work, we present a novel diagnosis toolkit, TracKlinic, for studying per factor tracking behaviors and exemplify its use on ten trackers. We show how TracKlinic helps identify potential challenge factors for a tracker. Our results suggest that, heavy shape variation and occlusion are worth more attentions in future research. Besides, we examine the ability of a tracker in handling different factors, which is crucial in understanding the strengths and weaknesses of a tracker. We investigate the failure rate of a tracker to each factor and observe that some rare factors such as out-of-view and occlusion with ration are fatal to visual trackers. We measure consistency of a tracker to different challenge factors and show that more efforts should be made to improve the stability of a tracker. 
By releasing TracKlinic, we aim to empower the community with more in-depth understanding of  tracking algorithms, beyond a single scalar metric evaluation. Most importantly, we expect that our diagnostic tool inspires the development of innovative tracking models that address the issues in current ones.

{\small
\bibliographystyle{ieee_fullname}
\bibliography{egbib}
}

\end{document}